\documentclass{article}

\PassOptionsToPackage{round}{natbib}

\usepackage[preprint]{neurips_2023}
\usepackage[utf8]{inputenc}
\usepackage[T1]{fontenc}
\usepackage{hyperref}
\usepackage{url}
\usepackage{booktabs}
\usepackage{amsfonts}
\usepackage{nicefrac}
\usepackage{microtype}
\usepackage{xcolor}
\usepackage{times}
\usepackage{latexsym}
\usepackage{multirow}
\usepackage{threeparttable}
\usepackage{xspace}
\usepackage{bm}
\usepackage{inconsolata}
\usepackage{booktabs}
\usepackage{pifont}
\usepackage{graphicx}
\usepackage{amsmath}
\usepackage{wrapfig}
\usepackage{multicol}
\usepackage{tabularx}
\usepackage[rightcaption]{sidecap}

\newcommand{\ie}{\emph{i.e.,}\xspace}

\title{Evaluating the External and Parametric Knowledge Fusion of Large Language Models}

\author{
  Hao Zhang\thanks{The first two authors contributed equally.},\; Yuyang Zhang\footnotemark[1],\; Xiaoguang Li,\; Wenxuan Shi,\; Haonan Xu,\; Huanshuo Liu\\
  \textbf{Yasheng Wang,\; Lifeng Shang,\; Qun Liu,\; Yong Liu,\; Ruiming Tang} \\
  Noah's Ark Lab, Huawei Technologies Co., Ltd\\
  \texttt{\{zhang.hao3,zhangyuyang4\}@huawei.com}\\
}

\begin{document}

\maketitle

\begin{abstract}
  Integrating external knowledge into large language models (LLMs) presents a promising solution to overcome the limitations imposed by their antiquated and static parametric memory. Prior studies, however, have tended to over-reliance on external knowledge, underestimating the valuable contributions of an LLMs' intrinsic parametric knowledge. The efficacy of LLMs in blending external and parametric knowledge remains largely unexplored, especially in cases where external knowledge is incomplete and necessitates supplementation by their parametric knowledge. We propose to deconstruct knowledge fusion into four distinct scenarios, offering the first thorough investigation of LLM behavior across each. We develop a systematic pipeline for data construction and knowledge infusion to simulate these fusion scenarios, facilitating a series of controlled experiments. Our investigation reveals that enhancing parametric knowledge within LLMs can significantly bolster their capability for knowledge integration. Nonetheless, we identify persistent challenges in memorizing and eliciting parametric knowledge, and determining parametric knowledge boundaries. Our findings aim to steer future explorations on harmonizing external and parametric knowledge within LLMs.
\end{abstract}

\section{Introduction}\label{sec:intro}
Parametric knowledge acquired by large language models (LLMs)~\citep{openai2023gpt4, touvron2023llama, anil2023palm2, du2022glm} during pre-training inevitably becomes outdated over time. Integrating additional contents into LLM inputs has emerged as an effective strategy to mitigate such issue~\citep{RAG—NEURIPS2020, nakano2021webgpt, retrieval_survey}. By incorporating external knowledge either into the input context~\citep{Retrieval-in-context, few_shot_rag} or through intermediary layers~\citep{retro, memorizing-transformer}, LLMs are endowed with more current information, expanding their knowledge boundary and reducing the instances of hallucinations and factual errors.

Many retrieval~\citep{RAG—NEURIPS2020, self-rag, few_shot_rag} or tool~\citep{shen2023hugginggpt, qin2024toolllm, schick2023toolformer} augmented methods predominantly rely on external evidence and often overlooking the rich knowledge stored within LLMs. Yet, the external evidence, inevitably, could be incomplete and noisy. While some approaches propose to refine the external evidence and post-calibrate the outputs by tapping into LLMs' parametric knowledge~\citep{k_edit, edit_survey}, the full potential of merging external with parametric knowledge remains unexplored. This paper aims to delve into \textit{how LLMs perform external and parametric knowledge fusion across various conditions}, especially when \textbf{LLMs encounter incomplete or irrelevant external knowledge}. A thorough understanding of this is crucial for a broader application of knowledge-augmented LLMs. Not only does this relate to the LLMs' parametric memory elicitation~\citep{xie2024adaptive, Qian2023MergeCE, Wang2023ResolvingKC, wang-etal-2023-self-knowledge}, but it is also associated with the knowledge boundary perception of LLMs~\citep{ren2023investigating, zhang2023merging, yin2023large}.

\begin{figure*}[t]
    \centering
    \includegraphics[trim={0cm 0cm 0cm 0.5cm},clip,width=0.98\textwidth]{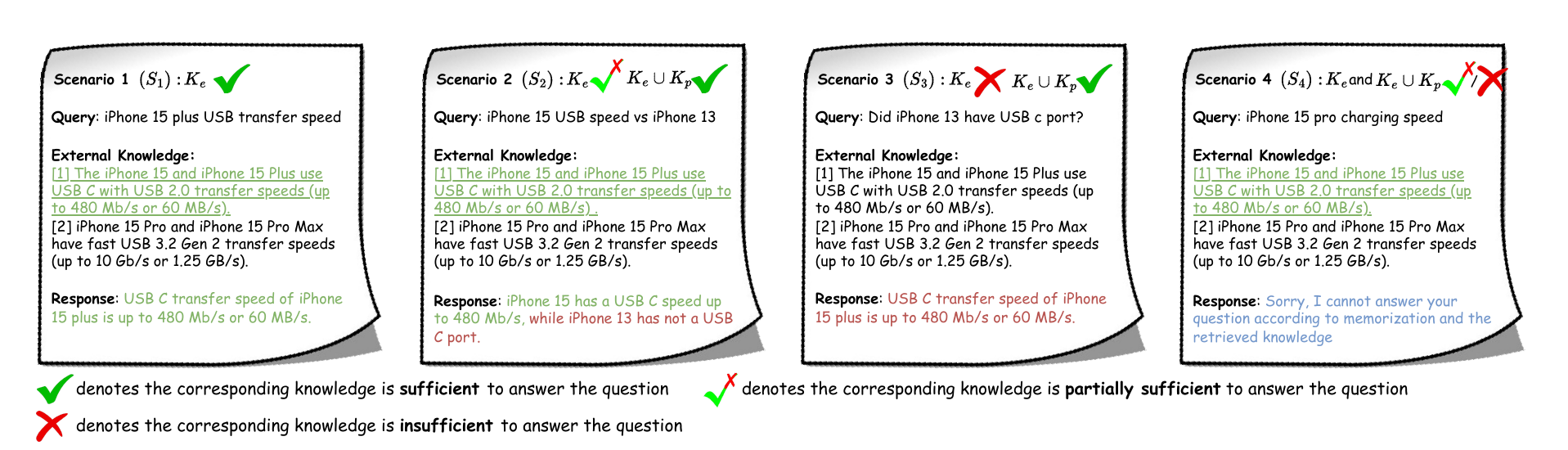}
    \vspace{-5pt}
    \caption{An illustration of four parametric and external knowledge fusion scenarios in LLMs.}
    \vspace{-5pt}
    \label{fig:intro_situations}
\end{figure*}

To elucidate the dynamics of LLMs in integrating external ($K_e$) and parametric ($K_p$) knowledge\footnote{For simplicity throughout this paper, we use $K_e$ and $K_p$ to symbolize the retrieved external knowledge and the LLMs' parametric knowledge, respectively.}, we define four distinct scenarios reflecting the interplay between $K_e$ and $K_p$ (depicted in Figure~\ref{fig:intro_situations}). The scenarios are as follows: (1) $S_1$ indicates that $K_e$ alone is sufficient to answer a query, independent of $K_p$'s contribution; (2) $S_2$ suggests that $K_e$ provides partial information, requiring $K_p$ to fill the gaps for a complete answer; (3) $S_3$ identifies situations where $K_e$ offers no useful information, and the answer depends solely on $K_p$; (4) $S_4$ describes cases where neither $K_e$ nor $K_p$ adequately address a query, making it theoretically unanswerable. Prior studies~\citep{robust_to_irrelevant, benchmarking_rag_chen} often presume situations where the availability of external knowledge ($K_e$) and $K_p$ is non-contributory, simplifying the knowledge fusion process to scenarios $S_1$ and $S_4$ and neglecting intermediate cases. The real challenge emerges when $K_e$ is sub-optimal, necessitating a nuanced integration of $K_e$ and $K_p$ for a cooperative response, especially in scenarios $S_2$ and $S_3$. However, the model-specific nature of LLMs' $K_p$ significantly complicates the precise delineation of knowledge boundaries and access to parametric knowledge. This complexity impedes a thorough and impartial evaluation of LLMs' capabilities in knowledge fusion.

To mitigate the challenges associated with acquiring parametric knowledge by LLMs, we propose a systematic pipeline for data construction and knowledge infusion. Specifically, we first collect the latest data from the electronic product domain and divide it into two parts: one for enhancing LLMs' parametric knowledge ($K_p$) through continued training, and the other as external knowledge ($K_e$). We also craft a set of questions based on the data to emulate the four scenarios: queries that solely depend on $K_e$ ($S_1$), queries requiring integration of $K_e$ and $K_p$ ($S_2$), queries dependent only on $K_p$ ($S_3$), and unanswerable queries ($S_4$). For each scenario, we provide relevant evidence and introduce additional distractors to mimic real-world conditions. Overall, this aims to standardize the parametric knowledge within different LLMs, facilitating equitable and model-independent evaluations.

We first inject new knowledge into LLMs through continued training and supervised fine-tuning, subsequently evaluating their knowledge retention. Then, we design a series of experiments to reveal the behaviors of LLMs in knowledge fusion. Despite the performance gains by integrating external and parametric knowledge, the results indicate that: (1) LLMs show deficiencies in recognizing domain knowledge, significantly influenced by their capacity to retain knowledge. (2) There are persistent challenges in memorizing and eliciting parametric knowledge and determining parametric knowledge boundaries for effective knowledge fusion. Our contributions are as follows:
\begin{list}{\labelitemi}{\leftmargin=1em}
\setlength{\topmargin}{0pt}
\setlength{\itemsep}{0em}
\setlength{\parskip}{0pt}
\setlength{\parsep}{0pt}
    \item We review knowledge fusion in LLMs, defining four distinct scenarios reflecting the interplay between external and parametric knowledge fusion for thorough evaluation.
    \item To mitigate the challenges associated with acquiring parametric knowledge by LLMs, we propose a systematic pipeline for data construction and knowledge infusion to facilitate knowledge fusion exploration.
    \item Through extensive experiments on various LLM backbones, we identify persistent challenges in memorizing and eliciting parametric knowledge and determining parametric knowledge boundaries. These challenges impair the effectiveness of knowledge fusion.
\end{list}

\section{Related Work}\label{sec:related_work}

\paragraph{Retrieval-augmented LLMs (RA-LLM).}
RA-LLM, including tools, are considered essential for linking LLMs with external knowledge sources~\citep{RAG—NEURIPS2020, qin2024toolllm, aug_llms_survey, retrieval_survey}, which makes LLMs more viable for practical applications. The prevalent methods either augment external evidence via in-context learning paradigm~\citep{lazaridou2022internet, he2022rethinking, few_shot_rag} or adopt external evidence to post-calibrate the generations~\citep{k_edit, li2024chainofknowledge, yan2024corrective}. Some work also suggests fine-tuning LLMs to enhance the utilization of external knowledge and optimize the retrieval strategy~\citep{RAG—NEURIPS2020, retro, self-rag, ra-dit}. These approaches mainly rely on external knowledge while overlooking the knowledge stored within LLMs, which may lead to undesirable results due to the biased and noisy external information~\citep{popqa, robust_to_irrelevant, conterfactual_k_ext}.

\paragraph{Parametric Knowledge in LLMs.} After pre-training, LLMs have internalized massive knowledge into their parameters, \textit{i.e.}, parametric knowledge~\citep{lm_as_kb, ffn_is_kv_memo, k_enhanced_survey, k_in_weight_space}. However, recent studies indicate that effectively leveraging LLMs' parametric knowledge is challenging~\citep{wang2023survey, Zhu2023PhysicsOL}. That is, although LLMs can memorize extensive knowledge, it does not guarantee their ability to adeptly elicit and manipulate it for subsequent tasks~\citep{reversal_curse, cannot_sove_ambiguity, fake_align, zhu2023physics_3_2}. Some work also observes that compared to directly augmenting knowledge into inputs, LLMs struggle to accurately memorize knowledge into parameters~\citep{llms_long_tail, sft-or-rag}. Besides, several studies explore the self-calibration~\citep{know_unknown_squad, mostly-know, may-not-know} and knowledge boundary detection~\citep{k_boudary} in LLMs, benefit for improving confidence and interpretability in their use of parametric knowledge, thus reducing hallucinations. Similarly, our objective is to investigate the capacity of LLMs for knowledge memorization and utilization in the knowledge fusion process, along with their self-calibration ability.

\paragraph{Knowledge Fusion of LLMs.}
To perform the fusion of external and parametric knowledge, \citet{jiang2023active} propose dynamically assessing the confidence level of model generation and intervening with retrieval at low confidence. \citet{wang-etal-2023-self-knowledge} elicit LLMs' ability to recognize their self-knowledge and achieve better knowledge integration. Some studies explore the knowledge conflict issues when integrating the external and parametric knowledge~\citep{control_robust, false_external, popqa, zhang-etal-2023-merging, xie2023adaptive}. However, these approaches mainly optimize knowledge fusion to enhance the subsequent tasks, such as open-domain QA~\citep{natural_qa, hotpotqa, strategyqa}, lacking a comprehensive evaluation of LLMs' behaviors in knowledge fusion. In contrast, we focus on the investigation of external and parametric knowledge fusion, including the systematic task definition, data construction pipeline, and thorough experiments.

\section{Task Definition}\label{sec:preliminary}
In practical applications, the external evidence obtained through retrieval or tools may be noisy, incomplete, or irrelevant~\cite{robust_to_irrelevant, conterfactual_k_ext}. This leads to the necessity of thoroughly considering various conditions when evaluating the external and parametric knowledge fusion. Therefore, we define four distinct scenarios capturing the diverse interactions between external and parametric knowledge of LLMs, aiming to encompass all potential circumstances as comprehensively as possible. Given external knowledge $K_e$ and parametric knowledge $K_p$, the defined scenarios are: (1) $S_1$ indicates $K_e$ alone is sufficient to answer a query, independent of $K_p$'s contribution; (2) $S_2$ denotes $K_e$ carries partial information, requiring $K_p$ to fill the gaps for a complete answer; (3) $S_3$ identifies situations where $K_e$ offers no useful information, and the answer depends solely on $K_p$; (4) $S_4$ describes cases where neither $K_e$ nor $K_p$ adequately address a query, making it theoretically unanswerable.

Suppose $K_p$ has been injected into the LLM. Formally, given a question $q_{S_i}$ and the corresponding external evidence $K_e^i$, where $i\in\{1, 2, 3, 4\}$, the response $\hat{a}_{S_i}$ of an LLM is generated as:
\begin{equation}
    \hat{a}_{S_i} = \texttt{LLM}_{(K_p)}([q_{S_i};K_e^i;\texttt{inst}]),
\end{equation}
where $\texttt{LLM}_{(K_p)}$ denotes the LLM already encoded the $K_p$ into its parameters, $\texttt{inst}$ represents the task-specific instructions. Ideally, for $S_1$, $K_e^1$ contains ground-truth evidence, where LLM can solely depend on $K_e^1$ to accurately answer $q_{s_1}$, that is, $\hat{a}_{S_1}$ is aligned with ground-truth $a_{S_1}$. For $S_2$, $K_e^2$ holds only partial information relevant to $q_{S_2}$, LLM also requires to elicit its corresponding $K_p$ to derive an accurate $\hat{a}_{S_2}$ for $q_{s_2}$. For $S_3$, $K_e^3$ is devoid of relevant information and is solely comprised of distractions, LLMs must eliminate these distractions and elicit its $K_p$ to reach the correct $\hat{a}_{S_3}$. In $S_4$, where $K_e^4$ consists solely of distractors and LLM lacks relevant $K_p$, it should opt to refrain from responding, implying that $\hat{a}_{S_4}$ should incorporate a refusal to answer $q_{S_4}$. Following the criteria outlined, we construct datasets, fine-tune different LLMs, and conduct a detailed evaluation of their ability to integrate external and parametric knowledge in these scenarios.

\section{Dataset Construction}\label{sec:data_collection}
Although LLMs encode massive knowledge through large-scale pre-training, the parametric knowledge of different LLMs exhibits notable variations due to discrepancies in training corpora, model scale, and forgetting issue~\citep{wang2023survey, llms_cl_survey}. Thus, it is challenging and almost infeasible to directly elicit the parametric knowledge of various LLMs~\citep{Qian2023MergeCE} and interact with external knowledge to conduct a fair and comprehensive evaluation. 

In this work, we focus on the assessment of knowledge fusion under a standard RAG setting. Facing difficulties in acquiring the parametric knowledge, we instead collect data to enrich LLMs' knowledge, enabling controlled and quantifiable evaluation of knowledge fusion. We split the data into two partitions, one part serving as external knowledge ($K_e$), and the other part integrated into the LLMs as parametric knowledge ($K_p$) through training. In this way, we eliminate the inconsistencies in $K_p$ among different LLMs. Leveraging the collected information, we further employ LLM to generate relevant question-answer pairs, forming a standard QA dataset for subsequent training and evaluation.

\begin{SCfigure}
	\centering
	\includegraphics[width=0.6\textwidth]{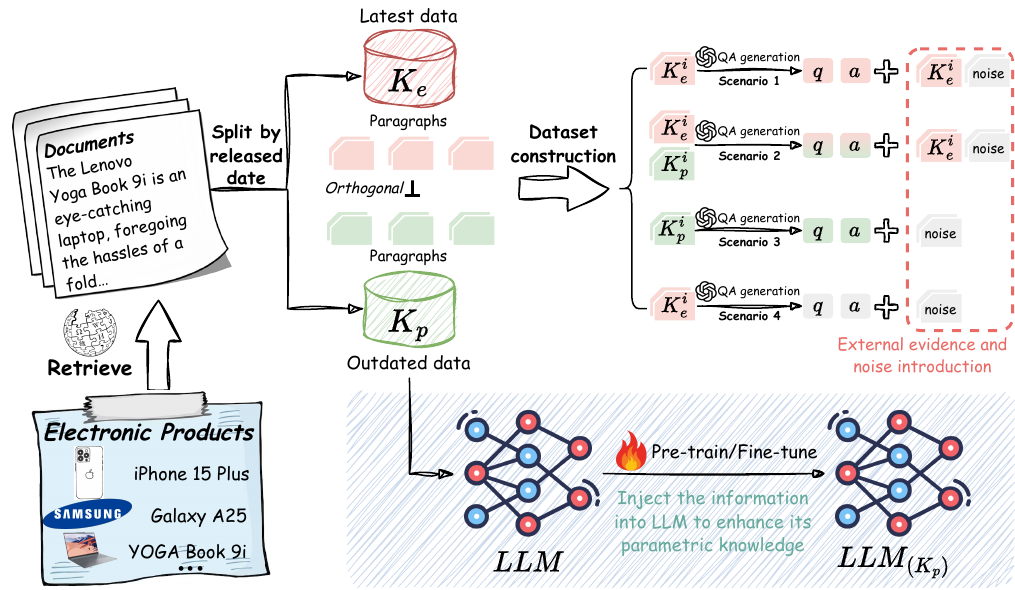}
	\caption{The overview of dataset construction. We first retrieve documents of electronics from websites. The documents are split into two portions based on their released date, and decomposed into paragraphs. The QA pairs for each scenario are generated via prompting LLMs, and the corresponding external evidence and noise are added as support sources. The outdated data is injected into LLMs through pre-training or fine-tuning.}
    \label{fig:fig2-workflow}
\end{SCfigure}

\begin{wrapfigure}{r}{0.4\textwidth} 
    \centering
    \includegraphics[width=0.4\textwidth]{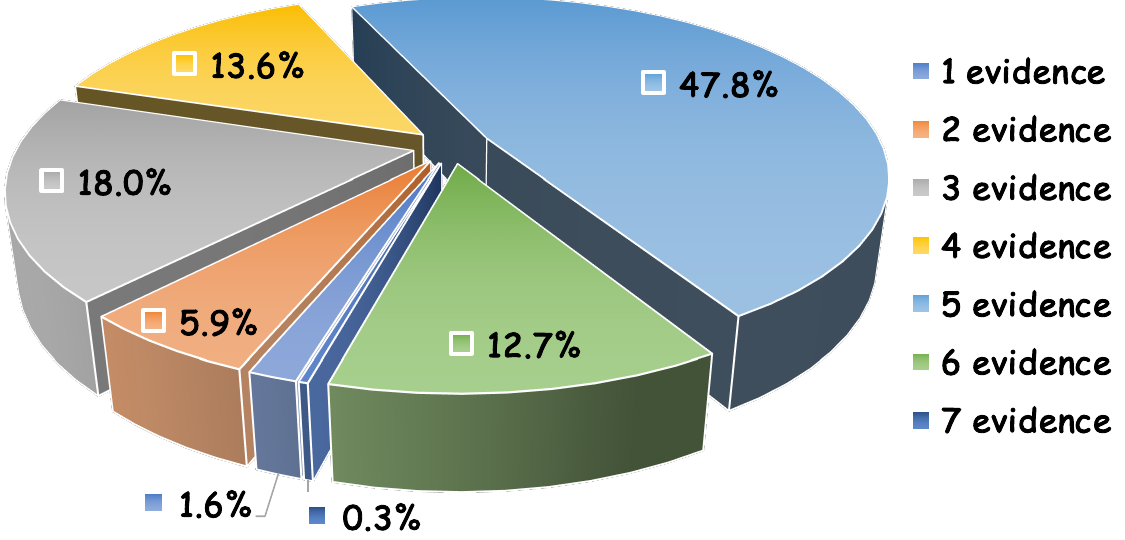}
    \caption{The distribution of the number of the associated evidence per QA sample.}
    \label{fig:each_sample}
\end{wrapfigure}

\paragraph{Data Source Preparation.} LLMs trained on cut-dated data are rarely exposed to domain-specific knowledge and have not yet encountered the latest information. Thus, the most recent and high-quality data sources are essential for building a viable dataset in our setting. Considering the swift evolution, variety, and annual surge of new products in the electronics domain, it emerges as a suitable source to measure knowledge fusion. Thus, we collect the data in the electronics domain spanning the preceding four years and utilize product introductory documents with detailed specifications serving as the primary source. Specifically, we collect over $500$ mobile phone names from websites and execute online searches to collate multiple search results for each product. Then, document filtration is applied through empirical rules and manual review, preserving documents with unique product introductions. Since some documents are too lengthy, we dissect them at the granularity of paragraphs and sentences to extract varied relevant information for each product. In general, we filtered out $1,700$ paragraphs from $5,000$ paragraphs in $1,500$ documents, where $900$ paragraphs are used to construct external knowledge, while $800$ paragraphs are trained in LLMs as parametric knowledge.

\paragraph{Dataset Construction.} The overview of dataset construction pipeline is shown in Figure~\ref{fig:fig2-workflow}. To simulate the external and parametric knowledge fusion scenarios, we divide the collected data into \textit{latest} and \textit{outdated} data according to their released date\footnote{The latest data is orthogonal to outdated data, \textit{i.e.}, no information overlap between them. Besides, the latest data refers to factual information about electronic products occurring after 2023-06-01, unseen by existing LLMs.}. We highlight that the outdated data may be learned by LLMs, while the latest data is less likely seen by LLMs due to cut-dated pre-training. We retain the \textit{latest data as external knowledge}, while the \textit{outdated data as parametric knowledge}, which is injected into LLM to enhance its parametric memory via fine-tuning. The evaluation of the LLMs' competence in fusing external and parametric knowledge focuses on its discriminating usage of external and parametric knowledge, as well as the correctness of the information utilized. To more clearly assess LLM capabilities, we use the partitioned latest and outdated data to develop respective QA evaluation datasets with sophisticated designed instructions. Specifically, we generate the QA pairs for four knowledge fusion scenarios as follows:
\begin{itemize}
    \setlength\itemsep{-0.2em}
    \item \textbf{Scenario 1} ($S_1$): We randomly select one or two snippets from the latest data to generate a QA pair with an LLM, and then the unrelated snippets are added as noise to the chosen snippets, creating the candidate knowledge for the generated QA pair.
    \item \textbf{Scenario 2} ($S_2$): We randomly select one snippet from both the latest and outdated data and generate a QA pair based on the two snippets. Unrelated snippets are added as noise to form the candidate knowledge for the QA pair.
    \item \textbf{Scenario 3} ($S_3$): The QA pair is generated solely based on the chosen snippets from outdated data, and noisy snippets are added to form candidate knowledge.
    \item \textbf{Scenario 4} ($S_4$): We randomly select one or two snippets from the latest data to create a QA pair, then discard the snippets and choose unrelated noise snippets as candidate knowledge for the QA pair, ensuring the unanswerability of the generated question.
\end{itemize}
To better align with real-world application settings, we adopt two noise introduction approaches to increase the challenge for LLMs in leveraging external knowledge. The first approach introduces noise snippets describing identical attributes across different electronic products, whereas the second approach presents noise snippets describing disparate attributes of a single electronic product. Moreover, to guarantee data quality, we further employ LLM evaluation coupled with manual review for data cleansing and filtering, yielding $210$, $580$, $200$, and $140$ samples per scenario.

\begin{wraptable}{r}{0.5\textwidth}
    \centering
    \begin{tabular}{c r r r r r}
    \toprule
    Data Split  & $S_1$   & $S_2$   & $S_3$   & $S_4$   & Total \\
    \midrule
    train & 100  & 390  & 90   & 50   & 630   \\
    dev   & 50   & 150  & 50   & 50   & 300   \\
    test  & 60   & 140  & 60   & 40   & 300   \\
    \bottomrule
    \end{tabular}
    \caption{The statistics of the dataset in each scenario.}
    \label{tab:dataset_split}
\end{wraptable}

\paragraph{Dataset Analysis.} We first assess the distribution of external evidence associated with each entity. Entities with three pieces of evidence are most prevalent, constituting $19.5\%$ of the sample, followed by entities with a single piece of evidence, accounting for $17.5\%$ of the total. We also examine the distribution of associated evidence per sample, depicted in Figure~\ref{fig:each_sample}. Notably, $47.8\%$ of samples contained five pieces of evidence, the highest proportion, while samples with three pieces of evidence comprised the second-highest percentage at $18\%$. Subsequently, we analyze the distribution of evidence lengths across the dataset, finding that quotes ranging from $500$ to $600$ characters represented the majority, totaling $78.6\%$. The overview of dataset partitioning is shown in Table~\ref{tab:dataset_split}. The dataset comprises training, validation, and test sets with $630$, $300$, and $300$ samples, respectively. The table also details the distribution of data across scenarios $S_1$, $S_2$, $S_3$, and $S_4$ within each subset.

\section{Experiment Setup}\label{sec:exp_setup}
\paragraph{Backbone Model.} We select the open-source ChatGLM3-6B~\citep{du2022glm} and Qwen-7B~\citep{Bai2023QwenTR}, and the black-box GPT-4~\citep{openai2023gpt4} as the backbones. These models are selected for their robust language understanding and instruction-following capabilities, which align well with our experiment design. Furthermore, the open-source LLMs enable flexible adaptation of model configurations and the analysis of their internal behaviors.

\paragraph{Parametric Knowledge Infusion.} For the selected LLMs, the outdated data portion needs to be injected into them via continued training or fine-tuning\footnote{Given the inaccessibility of GPT-4's weights, we assume it already memorizes outdated data and only conducts inference by providing external knowledge snippets as evidence. \texttt{GPT-4-0613} is used in all experiments.}. Since LLMs predominantly acquire knowledge in the pre-training phase, we also employ the same strategy for continued training. However, \citet{Zhu2023PhysicsOL} indicates that ``memorization of knowledge'' in language models merely means the model can fit the exact training data but does not imply it can extract the knowledge flexibly from data after training. To enhance knowledge memorization, we further adopt the data rewriting strategy suggested in \cite{Zhu2023PhysicsOL} to conduct data augmentation. Specifically, we use GPT-4~\citep{openai2023gpt4} to paraphrase the snippets in the outdated data portion and generate eight QA pairs related to that snippet as the supplementary data. The synthetic data is merged with the original data to train the backbones.

\paragraph{Evaluation Metrics.} We employ accuracy ($R_\text{acc}$) and information coverage ($R_\text{cover}$) as evaluation metrics to access the knowledge fusion capabilities of LLMs. Accuracy assesses if LLM responses accurately address the question and align with both external and parametric knowledge sources. Responses are deemed correct if consistent with these sources and incorrect if they include irrelevant content or deviate from the information provided. Information coverage refers to the degree to which LLMs encapsulate the core content of the reference. This coverage is classified into three categories: complete, partial, and no inclusion. Let $K_\text{gen}$ denote the knowledge contained in generations, $K_\text{gold}$ indicate the knowledge contained in the ground-truth answer, and $K_\text{ref}$ represents the given external and parametric knowledge\footnote{When evaluating LLMs' competence in $S_4$, we only measure their capability to decline to respond correctly.}. The $R_\text{acc}$ and $R_\text{cover}$ are computed as follows:

\begin{minipage}{.5\linewidth}
    \begin{equation}
      R_\text{acc} = \left\{
      \begin{matrix}
      1 & \text{if} \quad K_{\text{gen}} \subseteq K_{\text{ref}}\\
      0 & \text{otherwise}
    \end{matrix},
    \right.
    \end{equation}
  \end{minipage}%
  \begin{minipage}{.5\linewidth}
    \begin{equation}
       R_\text{cover} = \left\{
      \begin{matrix}
      \text{Complete} & \text{if} \quad K_{\text{gold}} \subseteq K_{\text{gen}}  \\
      \text{Partial} & \text{if} \quad K_{\text{gold}} \cap K_{\text{gen}} \\
      \text{Uncover} & \text{otherwise}
      \end{matrix}.
      \right.
    \end{equation}
\end{minipage}

\begin{table}[t]
    \centering
    \caption{Knowledge infusion results of ChatGLM~\cite{du2022glm} and Qwen~\cite{Bai2023QwenTR}.}
    \begin{tabular}{l c c c c }
    \toprule
    \multirow{2}{*}{Model} & \multirow{2}{*}{Accuracy (\%)} & \multicolumn{3}{c}{Coverage} \\
     \cmidrule(lr){3-5}
     & & Complete (\%) & Partial (\%) & Uncover (\%)  \\
    \midrule
    ChatGLM & 13.3 & 3.3 & 25.0 & 71.7 \\
    ChatGLM$_\text{CT}$ & 38.3 & 18.3 & 36.7 & 45.0 \\
    Qwen & 15.0 & 5.0 & 21.7 & 73.3 \\
    Qwen$_\text{CT}$ & 43.3 & 20.0 & 43.3 & 36.7 \\

    \bottomrule
    \end{tabular}
    \label{tab:inject_results}
\end{table}

\section{Experiment Results and Analysis}\label{sec:exp_results}
In this section, we conduct comprehensive experiments and in-depth analysis to investigate the knowledge fusion behaviors of various backbones. We use ChatGLM$_\text{CT}$ to represent ChatGLM that continues trained on $K_p$, and ChatGLM$_\text{CT\&SFT}$ denotes ChatGLM that continues trained on $K_p$ and further SFT on our train set. Similar to Qwen$_\text{CT}$ and Qwen$_\text{CT\&SFT}$. 

\subsection{The Performance of Knowledge Infusion}\label{ssec:k_inject_results}
To investigate the effectiveness of knowledge infusion by LLMs at continued training process, we conduct experiments using two models: ChatGLM3-6B (\textit{abbr.} ChatGLM) and Qwen-7B (\textit{abbr.} Qwen). These models are trained using the designated parametric knowledge partition, $K_p$. The evaluation involved querying the models with questions specifically related to $K_p$ to assess how well the models retained the trained knowledge. It is important to note that this evaluation is similar to scenario $S_3$, absent the inclusion of external knowledge distractors. We adopt the question-answer (QA) pairs from $S_3$ by excluding the associated external evidence to serve as our evaluation dataset.

The results, summarized in Table~\ref{tab:inject_results}, reveal that before continued training, both models demonstrated notably low accuracy rates: $13.3\%$ for ChatGLM and $15.0\%$ for Qwen, suggesting these models indeed have no such background knowledge. After continued training, there was a substantial enhancement in performance. Specifically, ChatGLM exhibits a $25\%$ absolute improvement in accuracy, while Qwen shows a $28.3\%$ absolute increase. This significant enhancement underscores the efficacy of continued training in injecting the knowledge into the models. Meanwhile, there is a notable enhancement in the model's ability to answer questions with complete and partial accuracy after continued training. Specifically, ChatGLM displays increases of $15\%$ and $11.7\%$ in complete and partial correct responses respectively, whereas Qwen showed improvements of $15\%$ and $21.6\%$.

Ideally, knowledge infusion through continued training of an LLM should enable the model to retain all imparted knowledge, resulting in the QA accuracy nearing $100\%$. In practice, however, even though accuracy significantly improves over untrained models, it remains considerably lower than the optimal situation. This suggests substantial amounts of knowledge are either not retained or not accurately elicited by the LLM. We highlight two key factors attributed to this issue: (i) model capability and (ii) dataset diversity. For the model capability, recent studies~\citep{Zhu2023PhysicsOL,zhu2023physics_3_2} highlight that LLM faces difficulties using its parametric knowledge, and processing such knowledge does not guarantee it to be elicited accurately. For the dataset diversity, the LLM simply memorizes the given knowledge, meaning it only fits the given contents and may not effectively utilize this knowledge. For instance, LLM is trained in a massive of documents during continued training and evaluated under the question-answering manner at test time, LLM may not effectively map the questions to the answers learned during training. Besides, altering the way questions are posed might prevent the LLM from providing correct answers, and the reversal curse~\citep{berglund2024the} is another example of such an issue. Thus, a straightforward solution is to diversify the given knowledge, such as constructing various types of QA pairs, paraphrasing the documents, etc., that training LLM to memorize the knowledge from different perspectives.

\begin{table}[t]
    \centering
    \caption{The overall performance of different LLMs under four scenarios. ``Direct'' represents directly prompting LLM to answer the questions by giving the corresponding external knowledge without continued training and supervised fine-tuning; ``SFT'' denotes the supervised fine-tuning on the train set of our constructed question-answering dataset; and ``CT'' means continuing training on the $K_p$ data partition to inject the knowledge into the LLM; ``Easy'' denotes the supervised fine-tuning on the train set as well as providing the supporting snippets during inference.}
    \begin{tabular}{c l c c c c c c c}
        \toprule
        \multirow{2}{*}{Scenario} & \multirow{2}{*}{Metric} & \multirow{2}{*}{GPT-4} & \multicolumn{4}{c}{ChatGLM} & \multicolumn{2}{c}{Qwen}  \\
        \cmidrule(lr){4-7} \cmidrule{8-9}
         & & & Direct & SFT & CT\&SFT & Easy & SFT & CT\&SFT \\
        \midrule
        \multirow{4}{*}{$S_1$} & $R_{\text{acc}}$ (\%) & 81.7 & 63.3 & 68.3 & 61.7 & 72.7 & 62.9 & 63.3 \\
        & Complete (\%) & 80.0 & 38.3 & 38.3 & 33.3 & 43.3 & 31.7 & 30.0 \\
        & Partial (\%) & 11.7 & 40.0 & 48.3 & 55.0 & 35.0 & 56.7 & 53.3 \\
        & Uncover (\%) & 8.3 & 21.7 & 13.4 & 11.7 & 21.7 & 11.6 & 16.7 \\
        \midrule
        \multirow{4}{*}{$S_2$} & $R_{\text{acc}}$ (\%) & 35.7 & 39.3 & 52.1 & 53.6 & 72.1 & 49.3 & 57.1 \\
        & Complete (\%) & 12.9 & 9.3 & 7.1 & 20.0 & 42.1 & 10.0 & 22.1 \\
        & Partial (\%) & 40.0 & 56.4 & 76.4 & 69.3 & 40.0 & 71.5 & 61.4 \\
        & Uncover (\%) & 47.1 & 34.3 & 16.5 & 10.7 & 17.9 & 18.5 & 16.5 \\
        \midrule
        \multirow{4}{*}{$S_3$} & $R_{\text{acc}}$ (\%) & 8.3 & 10.0 & 16.7 & 35.0 & 78.3 & 20.0 & 33.3 \\
        & Complete (\%) & 3.3 & 1.7 & 3.3 & 16.7 & 55.0 & 3.3 & 20.0 \\
        & Partial (\%) & 11.7 & 23.3 & 40.5 & 45.0 & 30.0 & 48.3 & 41.7 \\
        & Uncover (\%) & 85.0 & 75.0 & 56.2 & 38.3 & 15.0 & 48.4 & 38.3 \\
        \midrule
        $S_4$ & $R_{\text{acc}}$ (\%) & 37.5 & 25.0 & 30.0 & 40.0 & - & 27.5 & 40.0 \\
        \bottomrule
    \end{tabular}
    \label{tab:main_result}
\end{table}

\subsection{Main Results}\label{ssec:main_results}
In this section, we conduct a comprehensive evaluation of different LLMs over the four knowledge fusion scenarios. The results are summarized in Table~\ref{tab:main_result}. Note ``Direct'' mode denotes that we directly prompt LLM to answer the questions by giving the corresponding external knowledge without continued training or supervised fine-tuning; ``SFT'' mode represents that we supervised fine-tune our constructed train set; ``CT\&SFT'' mode denotes that we continue training the LLM on $K_p$ following by further supervised fine-tune on the constructed train set; and ``Easy'' mode means that we not only SFT the LLM on the constructed train set but also provide the ground-truth snippets coupled with distractors during inference.

\subsubsection{Knowledge Fusion Performance on $S_1$}
Scenario 1, $S_1$, denotes that provided external knowledge, $K_e$, alone is sufficient to answer a question, independent of $K_p$'s contribution. The $S_1$ results of the different models are summarized in Table~\ref{tab:main_result}. Observed that GPT-4 achieves the best performance among all models, which obtains $81.7\%$ accuracy and $66.7\%$ complete coverage. The higher accuracy usually leads to better ``complete'' and/or ``partial'' coverage. Compared to ChatGLM and Qwen, GPT-4 has a richer internal knowledge base and more powerful content comprehension capabilities. For ChatGLM, ChatGLM$_{\text{SFT}}$ is superior to ChatGLM$_{\text{Direct}}$, \ie vanilla ChatGLM without continued train or SFT, by $5\%$ absolute improvements on accuracy. Notably, the continued training does not always contribute to the performance improvements in scenario 1 ($S_1$). For instance, ChatGLM$_{\text{SFT}}$ obtains $68.3\%$ accuracy while ChatGLM$_{\text{CT\&SFT}}$ only achieves $61.7\%$. Qwen$_{\text{SFT}}$ is comparable to the Qwen$_{\text{CT\&SFT}}$ with only $0.4\%$ accuracy gap. We highlight it is because all the ground-truth evidence is provided by external knowledge in $S_1$, SFT helps LLM to learn how to follow the instructions and utilize the given knowledge to reach the correct responses. The knowledge provided by CT is useless in the $S_1$ scenario, and continued training may inevitably lead to capability degradation of LLM~\citep{shi2024continual}. Nevertheless, ChatGLM$_{\text{SFT}}$ is inferior to ChatGLM$_{\text{Easy}}$ with a distinct gap, \ie $68.3\%$ versus $72.2\%$, which demonstrates that noisy external knowledge indeed affects LLMs adversely~\citep{pan2024contexts,cuconasu2024power}. Notably, the noise in our dataset is carefully curated, being relevant yet useless for effective responses.

\subsubsection{Knowledge Fusion Performance on $S_2$}\label{sssec:s_2_results}
Scenario 2, $S_2$, represents that $K_e$ provides \textit{partial} knowledge to answer a question, and it requires $K_p$ to fill the gaps for a complete answer. As summarized in Table~\ref{tab:main_result}, the overall performance of different backbone models in $S_2$ is significantly inferior to that in $S_1$. For instance, although the test cases are different, the accuracy of GPT-4 drops from $81.7\%$ to $35.7\%$ and the complete coverage drops from $80.0\%$ to $12.9\%$, which proves that our data partition, \ie $K_e$ and $K_p$, is reasonable, indicating that the knowledge in $K_p$ is not covered by the off-the-shelf LLMs. Compared to ChatGLM$_{\text{Direct}}$, ChatGLM$_{\text{SFT}}$ achieve much better performance, $52.1\%$ versus $39.3\%$. Since SFT does not inject $K_p$ into LLM, both ChatGLM$_{\text{Direct}}$ and ChatGLM$_{\text{SFT}}$ suffer from low complete coverage. Comparing ChatGLM$_{\text{SFT}}$ with ChatGLM$_{\text{CT\&SFT}}$, the complete coverage of ChatGLM$_{\text{CT\&SFT}}$ is higher than ChatGLM$_{\text{SFT}}$ by a large margin, which indicates that CT indeed injects the $K_p$ into the LLMs, the LLMs are capable of using the knowledge to answer the questions by considering both its parametric knowledge and the given external knowledge. A similar observation is held for Qwen model. 

However, the accuracy of CT\&SFT is slightly better than that of SFT for both ChatGLM and Qwen models. We emphasize that there are two aspects to this issue. One primary factor is the model's memory capacity, which determines the extent of knowledge retained during training. As discussed in Section~\ref{ssec:k_inject_results}, due to limitations in model capacity and dataset diversity, the model can accurately retain only a subset of the provided $K_p$. Another factor is that LLMs face difficulties using their parametric knowledge~\citep{Zhu2023PhysicsOL,zhu2023physics_3_2} and accurate parametric and external knowledge fusion for question answering is challenging. According to case studies in $S_2$, we observe that the success rate of LLM's parametric knowledge elicitation is only around $60\%$. If we directly use all the ground-truth supporting snippets as external knowledge and feed them into LLMs, the LLMs' performance increases significantly (see ``Easy'' and ``CT\&SFT''), which further proves the deficiency of LLMs to utilize their parametric knowledge.
Some work~\citep{jeong2024adaptiverag,ding2024survey} performs parametric and external knowledge fusion by first producing partial answers using parametric knowledge and then integrating the generated knowledge and external knowledge for final answer generation. In contrast, we directly prompt LLM to generate the final answer by considering its parametric knowledge and the given external knowledge.

\subsubsection{Knowledge Fusion Performance on $S_3$}
Note that scenario 3, $S_3$, simulates the situation that $K_e$ offers no useful information and the correct answer depends solely on $K_p$. As reported in Table~\ref{tab:main_result}, without SFT or CT, all the evaluated backbone models fail in $S_3$. For instance, GPT-4 and ChatGLM$_{\text{Direct}}$ only obtain $8.3\%$ and $10.0\%$ accuracy, respectively. Compared to ChatGLM$_{\text{Direct}}$, ChatGLM$_{\text{SFT}}$ slightly improves the performance, since SFT only teaches LLM to follow the instruction for answer generation, while it does not inject the new knowledge into the LLM. After injecting the $K_p$ into LLM, ChatGLM$_{\text{CT\&SFT}}$ significantly outperforms ChatGLM$_{\text{SFT}}$ in accuracy by $18.3\%$ absolute improvement. A similar result is observed for the Qwen model, which obtains $13.3\%$ absolute gains. Despite the improvements achieved, their performance is still sub-optimal. For instance, ChatGLM$_{\text{Easy}}$ reached $78.3\%$ accuracy in $S_3$, which $43.3\%$ higher than ChatGLM$_{\text{CT\&SFT}}$. Similar to the observation in Section~\ref{sssec:s_2_results}, the results indicate that CT cannot guarantee that LLM will fully retain all knowledge, and LLM itself faces difficulties in accurately eliciting parametric knowledge. Moreover, in the knowledge infusion experiment (ref. Section~\ref{ssec:k_inject_results}), we use the same QA pairs as $S_3$, but we ignore all the external distractors. Comparing the results between knowledge infusion (see Table~\ref{tab:inject_results}) and $S_3$ knowledge fusion (see Table~\ref{tab:main_result}), we observe that the accuracies of both ChatGLM and Qwen in $S_3$ knowledge fusion are lower than that in knowledge infusion, which emphasizes that incorporating noisy external knowledge negatively impacts LLM performance, as it may cause overconfidence in plausible but incorrect information.

\subsubsection{Knowledge Fusion Performance on $S_4$}
Recall that scenario 4, $S_4$, describes cases where neither $K_e$ nor $K_p$ adequately address the questions, making those questions theoretically unanswerable. This scenario aims to evaluate the efficacy of LLMs to correctly provide a refusal response if they do not have the corresponding parametric knowledge and the external knowledge is unhelpful. As reported in Table~\ref{tab:main_result}, all the evaluated backbone models, including GPT-4, fail to trigger the refusal response under $S_4$. In general, these models tend to be overconfident in the provided plausible but incorrect external knowledge, yielding wrong answers. SFT shows positive impacts on performance improvement, where ChatGLM$_{\text{SFT}}$ is $5\%$ higher than ChatGLM$_{\text{Direct}}$. Due to the presence of some refusal response samples in the data, SFT can guide the LLM on how to trigger and issue refusals to some extent. Comparing ChatGLM$_{\text{CT\&SFT}}$ with ChatGLM$_{\text{SFT}}$, CT further boosts the performance. We speculate that continued training with domain knowledge improves the LLM's field-specific understanding, enhancing its ability to discern whether the provided external knowledge and its parametric knowledge can effectively address a given question.

\subsection{Findings and Challenges}
According to the in-depth analyses presented in Section~\ref{ssec:k_inject_results} and~\ref{ssec:main_results}, we conclude the observations and insights as follows:

\begin{itemize}
    \item \textbf{Noise Robustness of LLM}: Noise and interference information from external knowledge negatively impact LLM performance~\citep{chen2024benchmarking}, as evidenced across multiple LLMs in scenarios $S_1\sim S_4$, leading to the generation of seemingly plausible but incorrect answers.
    \item \textbf{Impact of supervised fine-tuning}: Across all scenarios, $S_1\sim S_4$, supervised fine-tuning (SFT) helps to improve the performance of LLMs. Despite SFT (almost) does not inject new knowledge into the LLM, it enhances the LLM's ability in instruction adherence, leading to more standardized outcomes~\citep{Zhu2023PhysicsOL,zhu2023physics_3_2}.
    \item \textbf{Impact of continued training}: when external knowledge is sufficient, \ie $S_1$, domain knowledge infusion via continued training yields negligible improvement, as the LLM can generate correct answers based solely on the provided information. Conversely, when external knowledge is inadequate, \ie $S_2\sim S_4$, continued training is crucial and significantly enhances performance~\citep{jiao2023panda,naveed2024comprehensive,fujii2024continual}, since LLM lacks the necessary domain knowledge and continued training can effectively alleviate the knowledge limitations of LLM.
    \item \textbf{The effect of knowledge infusion}: Although experiments on knowledge infusion and $S_2\sim S_3$ demonstrate the effectiveness, performance gains of knowledge infusion remain limited. Due to constraints in model capacity and dataset diversity, LLMs can retain only a subset of the knowledge accurately via continued training~\citep{moiseev2022skill,arrotta2024contextgpt}. Additionally, LLMs also struggle to utilize parametric knowledge effectively, and processing such knowledge does not ensure accurate elicitation~\citep{zhu2023physics_3_2}.
    \item \textbf{The effect of refusal}: Ideally, LLM should issue a refusal response when external knowledge is irrelevant and lacks corresponding parametric knowledge. However, LLMs tend to be overconfident in external knowledge regardless of its usefulness~\citep{chen2024benchmarking}, particularly for cases like $S_3 \sim S_4$ where external knowledge is entirely unhelpful, leading to plausible but incorrect responses (hallucinations). SFT and CT ameliorate this issue. SFT provides examples of refusal responses in the training data, instructing the LLM when to refuse. Meanwhile, CT enhances the LLM's understanding of domain knowledge, improving its ability to judge the efficacy of both external and parametric knowledge in addressing a given question.
    \item \textbf{The effect of knowledge fusion}: When the external knowledge is incomplete, LLMs often struggle to effectively fuse parametric and external information for response generation~\citep{xie2023adaptive}. Efficient fusion is generally constrained by factors such as the LLM's knowledge capacity, knowledge boundary perception, noise resistance, and knowledge elicitation ability~\citep{wang-etal-2023-self-knowledge}.
\end{itemize}

Accordingly, to better fuse parametric and external knowledge in LLMs, we identify several key challenges that need addressing. While some work~\citep{Zhu2023PhysicsOL,zhu2023physics_3_2,chen2024benchmarking,wang-etal-2023-self-knowledge,xie2023adaptive} are underway to tackle these issues, the approach from the perspective of knowledge fusion remains underexplored.
\begin{itemize}
    \item With respect to the noisy information, how to eliminate noise in external knowledge and enhance the noise resistance ability of LLMs, especially in the absence of corresponding parametric knowledge?
    \item For knowledge infusion, how to optimize the training strategies or methodologies so that the LLM can retain as much knowledge as possible?
    \item How can LLMs elicit the correct parametric knowledge to answer given questions and accurately recognize its knowledge boundaries, triggering a refusal when neither parametric nor external knowledge is available, rather than generating a hallucinated response?
    \item How can we optimize the use of parametric and external knowledge to achieve accurate integration when external knowledge is incomplete and the LLM has corresponding default knowledge?
\end{itemize}

\section{Conclusion}
This work underscores the nuanced interplay between external and parametric knowledge within LLMs, emphasizing the potential and challenges intrinsic to their fusion. By meticulously deconstructing knowledge fusion into four distinct scenarios and developing a structured pipeline for data construction and knowledge infusion, we have provided a comprehensive examination of LLM behavior across varying contexts of knowledge supplementation. The results indicate that while supervised fine-tuning or enhancing parametric knowledge via continued training is capable of improving the knowledge fusion performance, persistent challenges remain in noise resistance, more effective knowledge infusion, parametric knowledge boundary perception, and accurate knowledge elicitation. These insights lay a foundational framework for future research aimed at achieving a more harmonious and effective synthesis of external and parametric knowledge within LLMs, ultimately advancing their capabilities and applications.

\bibliographystyle{plainnat}
\bibliography{neurips_2023}

\end{document}